# Gradient Magnitude Similarity Deviation: An Highly Efficient Perceptual Image Quality Index

Wufeng Xue, Lei Zhang, *Member IEEE*, Xuanqin Mou, *Member IEEE*, and Alan C. Bovik, *Fellow, IEEE*

***Abstract*—It is an important task to faithfully evaluate the perceptual quality of output images in many applications such as image compression, image restoration and multimedia streaming. A good image quality assessment (IQA) model should not only deliver high quality prediction accuracy but also be computationally efficient. The efficiency of IQA metrics is becoming particularly important due to the increasing proliferation of high-volume visual data in high-speed networks. We present a new effective and efficient IQA model, called gradient magnitude similarity deviation (GMSD). The image gradients are sensitive to image distortions, while different local structures in a distorted image suffer different degrees of degradations. This motivates us to explore the use of global variation of gradient based local quality map for overall image quality prediction. We find that the pixel-wise gradient magnitude similarity (GMS) between the reference and distorted images combined with a novel pooling strategy – the standard deviation of the GMS map – can predict accurately perceptual image quality. The resulting GMSD algorithm is much faster than most state-of-the-art IQA methods, and delivers highly competitive prediction accuracy. MATLAB source code of GMSD can be downloaded at http://www4.comp.polyu.edu.hk/~cslzhang/IQA/GMSD/GMSD.htm.**

## I. Introduction

It is an indispensable step to evaluate the quality of output images in many image processing applications such as image acquisition, compression, restoration, transmission, etc. Since human beings are the ultimate observers of the processed images and thus the judges of image quality, it is highly desired to develop automatic approaches that can predict perceptual image quality consistently with human subjective evaluation. The traditional mean square error (MSE) or peak signal to noise ratio (PSNR) correlates poorly with human perception, and hence researchers have been devoting much effort in developing advanced perception-driven image quality assessment (IQA) models [2, 25]. IQA models can be classified [3] into full reference (FR) ones, where the pristine reference image is available, no reference ones, where the reference image is not available, and reduced reference ones, where partial information of the reference image is available.

Wufeng Xue and Xuanqin Mou are with the Institute of Image Processing and Pattern Recognition, Xi'an Jiaotong University, Xi'an, 710049, China (e-mail: xwolfs@hotmail.com, xqmou@mail.xjtu.edu.cn).

Lei Zhang is with the Dept. of Computing, The Hong Kong Polytechnic University, Hong Kong, China. (e-mail: cslzhang@comp.polyu.edu.hk)

Alan C. Bovik is with the with the Laboratory for Image and Video Engineering (LIVE), Dept. of Electrical and Computer Engineering, The University of Texas at Austin, Austin, TX 78712 USA. (e-mail: bovik@ece.utexas.edu)

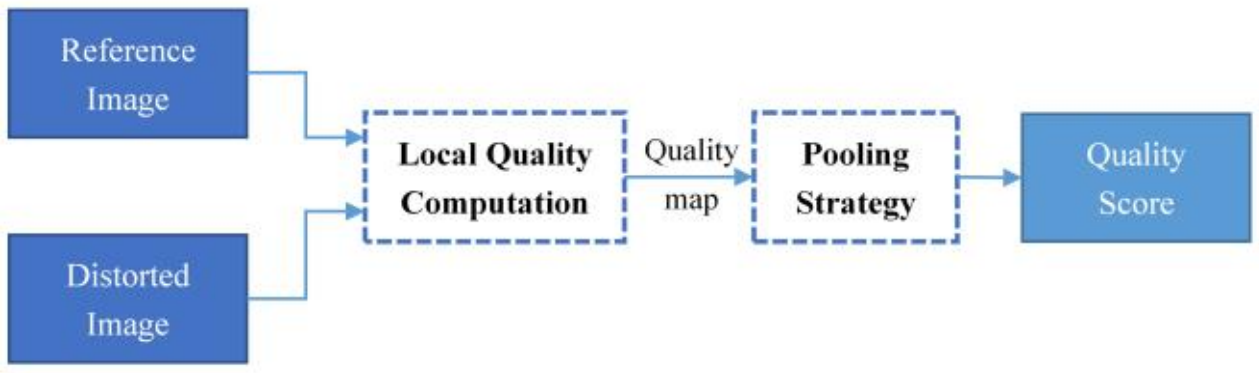

**Figure 1** The flowchart of a class of two-step FR-IQA models.

This paper focuses on FR-IQA models, which are widely used to evaluate image processing algorithms by measuring the quality of their output images. A good FR-IQA model can shape many image processing algorithms, as well as their implementations and optimization procedures [1]. Generally speaking, there are two strategies for FR-IQA model design. The first strategy follows a bottom-up framework [3, 30], which simulates the various processing stages in the visual pathway of human visual system (HVS), including visual masking effect [32], contrast sensitivity [33], just noticeable differences [34], etc. However, HVS is too complex and our current knowledge about it is far from enough to construct an accurate bottom-up IQA framework. The second strategy adopts a top-down framework [3, 30, 4-8], which aims to model the overall function of HVS based on some global assumptions on it. Many FR-IQA models follow this framework. The well-known Structure SIMilarity (SSIM) index [8] and its variants, Multi-Scale SSIM (MS-SSIM) [17] and Information Weighted SSIM (IW-SSIM) [16], assume that HVS tends to perceive the local structures in an image when evaluating its quality. The Visual Information Fidelity (VIF) [23] and Information Fidelity Criteria (IFC) [22] treat HVS as a communication channel and they predict the subjective image quality by computing how much the information within the perceived reference image is preserved in the perceived distorted one. Other state-of-the-art FR-IQA models that follow the top-down framework include Ratio of Non-shift Edges (rNSE) [18, 24], Feature SIMilarity (FSIM) [7], etc. A comprehensive survey and comparison of state-of-the-art IQA models can be found in [30, 14].

Aside from the two different strategies for FR-IQA model design, many IQA models share a common two-step framework [16, 4-8], as illustrated in Fig. 1. First, a *local quality map* (LQM) is computed by locally comparing the distorted image with the reference image via some similarity function. Then a single overall quality score is computed from the LQM via some *pooling* strategy. The simplest and widely used pooling strategy is average pooling, i.e., taking the average of local quality values as the overall quality prediction score. Since different regions may contribute differently to the overall perception of an image's quality, the local quality values can be weighted to produce the final quality score. Example weighting strategies include local measures of information content [9, 16], content-based partitioning [19], assumed visual fixation [20], visual attention [10] and distortion based weighting [9, 10, 29]. Compared with average pooling, weighted pooling can improve the IQA accuracy to some extent; however, it may be costly to compute the weights. Moreover, weighted pooling complicates the pooling process and can make the predicted quality scores more nonlinear w.r.t. the subjective quality scores (as shown in Fig. 5).

In practice, an IQA model should be not only effective (i.e., having high quality prediction accuracy) but also efficient (i.e., having low computational complexity). With the increasing ubiquity of digital imaging and communication technologies in our daily life, there is an increasing vast amount of visual data to be evaluated. Therefore, efficiency has become a critical issue of IQA algorithms. Unfortunately, effectiveness and efficiency are hard to achieve simultaneously, and most previous IQA algorithms can reach only one of the two goals. Towards contributing to filling this need, in this paper we develop an efficient FR-IQA model, called gradient magnitude similarity deviation (GMSD). GMSD computes the LQM by comparing the gradient magnitude maps of the reference and distorted images, and uses standard deviation as the pooling strategy to compute the final quality score. The proposed GMSD is much faster than most state-of-the-art FR-IQA methods, but supplies surprisingly competitive quality prediction performance.

Using image gradient to design IQA models is not new. The image gradient is a popular feature in IQA [4-7, 15, 19] since it can effectively capture image local structures, to which the HVS is highly sensitive. The most commonly encountered image distortions, including noise corruption, blur and compression artifacts, will lead to highly visible structural changes that "pop out" of the gradient domain. Most gradient based FR-IQA models [5-7, 15] were inspired by SSIM [8]. They first compute the similarity between the gradients of reference and distorted images, and then compute some additional information, such as the difference of gradient orientation, luminance similarity and phase congruency similarity, to combine with the gradient similarity for pooling. However, the computation of such additional information can be expensive and often yields small performance improvement.

Without using any additional information, we find that using the image gradient magnitude alone can still yield highly accurate quality prediction. The image gradient magnitude is responsive to artifacts introduced by compression, blur or additive noise, etc. (Please refer to Fig. 2 for some examples.) In the proposed GMSD model, the pixel-wise similarity between the gradient magnitude maps of reference and distorted images is computed as the LQM of the distorted image. Natural images usually have diverse local structures, and different structures suffer different degradations in gradient magnitude. Based on the idea that the global variation of local quality degradation can reflect the image quality, we propose to compute the standard deviation of the gradient magnitude similarity induced LQM to predict the overall image quality score. The proposed standard deviation pooling based GMSD model leads to higher accuracy than all state-of-the-art IQA metrics we can find, and it is very efficient, making large scale real time IQA possible.

The rest of the paper is organized as follows. Section II presents the development of GMSD in detail. Section III presents extensive experimental results, discussions and computational complexity analysis of the proposed GMSD model. Finally, Section IV concludes the paper.

## II. Gradient Magnitude Similarity Deviation

### A. Gradient Magnitude Similarity

The image gradient has been employed for FR-IQA in different ways [3, 4, 5, 6, 7, 15]. Most gradient based FR-IQA methods adopt a similarity function which is similar to that in SSIM [8] to compute gradient similarity. In SSIM, three types of similarities are computed: luminance similarity (LS), contrast similarity (CS) and structural similarity (SS). The product of the three similarities is used to predict the image local quality at a position. Inspired by SSIM, Chen *et al.* proposed gradient SSIM (G-SSIM) [6]. They retained the LS term of SSIM but applied the CS and SS similarities to the gradient magnitude maps of reference image (denoted by $\mathbf{r}$) and distorted image (denoted by $\mathbf{d}$). As in SSIM, average pooling is used in G-SSIM to yield the final quality score. Cheng *et al.* [5] proposed a geometric structure distortion (GSD) metric to predict image quality, which computes the similarity between the gradient magnitude maps, the gradient orientation maps and contrasts of $\mathbf{r}$ and $\mathbf{d}$. Average pooling is also used in GSD. Liu *et al.* [15] also followed the framework of SSIM. They predicted the image quality using a weighted summation (i.e., a weighted pooling strategy is used) of the squared luminance difference and the gradient similarity. Zhang *et al.* [7] combined the similarities of phase congruency maps and gradient magnitude maps between $\mathbf{r}$ and $\mathbf{d}$. A phase congruency based weighted pooling method is used to produce the final quality score. The resulting Feature SIMilarity (FSIM) model is among the leading FR-IQA models in term of prediction accuracy. However, the computation of phase congruency features is very costly.

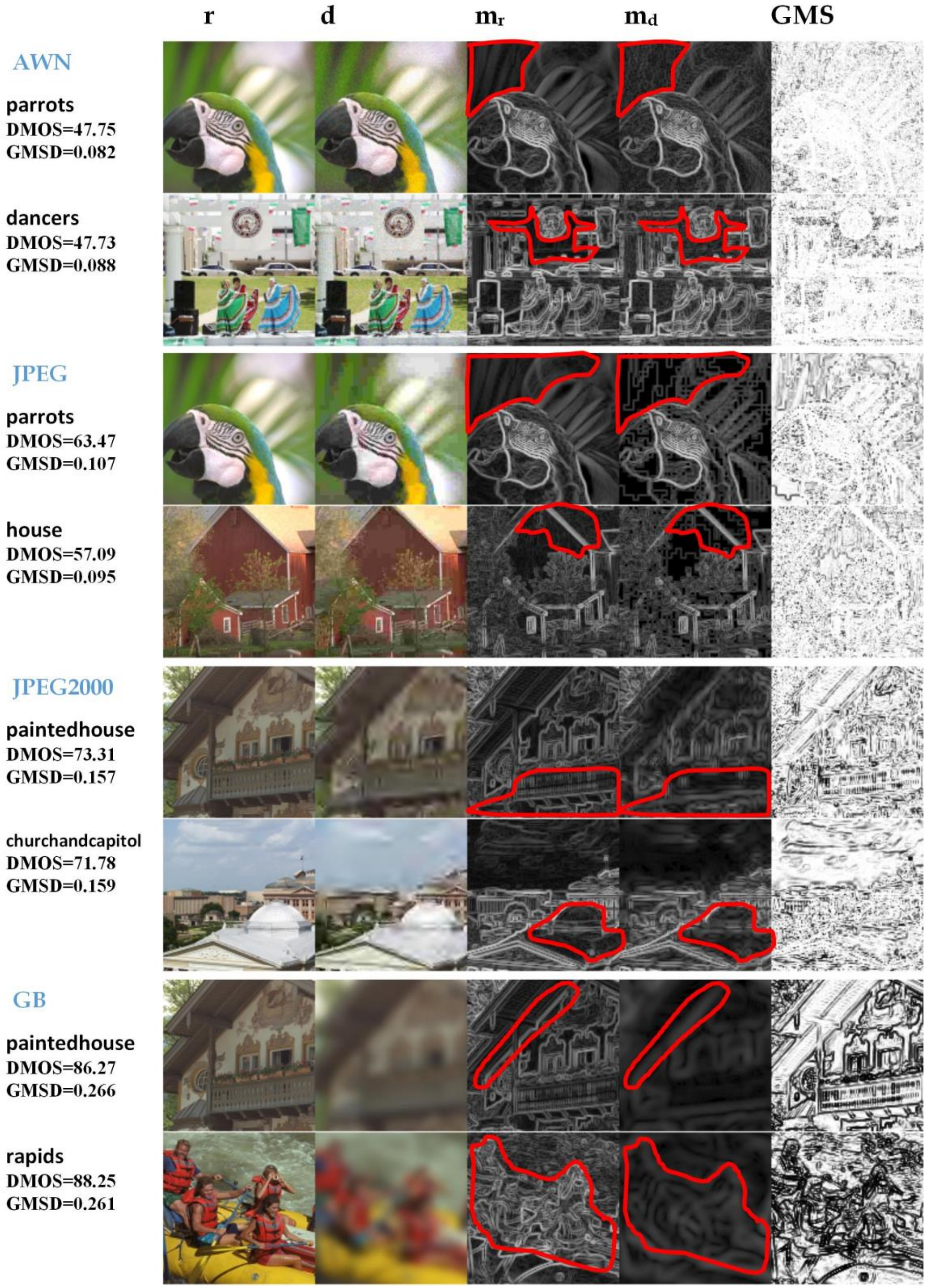

**Figure 2** Examples of reference (**r**) and distorted (**d**) images, their gradient magnitude images ($\mathbf{m}_r$ and $\mathbf{m}_d$), and the associated gradient magnitude similarity (GMS) maps, where brighter gray level means higher similarity. The highlighted regions (by red curve) are with clear structural degradation in the gradient magnitude domain. From top to bottom, the four types of distortions are additive white noise (AWN), JPEG compression, JPEG2000 compression, and Gaussian blur (GB). For each type of distortion, two images with different contents are selected from the LIVE database [11]. For each distorted image, its subjective quality score (DMOS) and GMSD index are listed. Note that distorted images with similar DMOS scores have similar GMSD indices, though their contents are totally different.

For digital images, the gradient magnitude is defined as the root mean square of image directional gradients along two orthogonal directions. The gradient is usually computed by convolving an image with a linear filter such as the classic Roberts, Sobel, Scharr and Prewitt filters or some task-specific ones [26, 27, 28]. For simplicity of computation and to introduce a modicum of noise-insensitivity, we utilize the Prewitt filter to calculate the gradient because it is the simplest one among the 3×3 template gradient filters. By using other filters such as the Sobel and Scharr filters, the proposed method will have similar IQA results. The Prewitt filters along horizontal ($x$) and vertical ($y$) directions are defined as:

$$\mathbf{h}_x = \begin{bmatrix} 1/3 & 0 & -1/3 \\ 1/3 & 0 & -1/3 \\ 1/3 & 0 & -1/3 \end{bmatrix},\ \mathbf{h}_y = \begin{bmatrix} 1/3 & 1/3 & 1/3 \\ 0 & 0 & 0 \\ -1/3 & -1/3 & -1/3 \end{bmatrix} \quad (1)$$

Convolving $\mathbf{h}_x$ and $\mathbf{h}_y$ with the reference and distorted images yields the horizontal and vertical gradient images of $\mathbf{r}$ and $\mathbf{d}$. The gradient magnitudes of $\mathbf{r}$ and $\mathbf{d}$ at location $i$, denoted by $\mathbf{m}_r(i)$ and $\mathbf{m}_d(i)$, are computed as follows:

$$\mathbf{m}_r(i) = \sqrt{(\mathbf{r} \otimes \mathbf{h}_x)^2(i) + (\mathbf{r} \otimes \mathbf{h}_y)^2(i)} \quad (2)$$

$$\mathbf{m}_d(i) = \sqrt{(\mathbf{d} \otimes \mathbf{h}_x)^2(i) + (\mathbf{d} \otimes \mathbf{h}_y)^2(i)} \quad (3)$$

where symbol “ $\otimes$ ” denotes the convolution operation.

With the gradient magnitude images $\mathbf{m}_r$ and $\mathbf{m}_d$ in hand, the gradient magnitude similarity (GMS) map is computed as follows:

$$GMS(i) = \frac{2\mathbf{m}_r(i)\mathbf{m}_d(i) + c}{\mathbf{m}_r^2(i) + \mathbf{m}_d^2(i) + c} \quad (4)$$

where $c$ is a positive constant that supplies numerical stability, $L$ is the range of the image intensity. (The selection of $c$ will be discussed in Section III-B.) The GMS map is computed in a pixel-wise manner; nonetheless, please note that a value $\mathbf{m}_r(i)$ or $\mathbf{m}_d(i)$ in the gradient magnitude image is computed from a small local patch in the original image $\mathbf{r}$ or $\mathbf{d}$.

The GMS map serves as the local quality map (LQM) of the distorted image $\mathbf{d}$. Clearly, if $\mathbf{m}_r(i)$ and $\mathbf{m}_d(i)$ are the same, $GMS(i)$ will achieve the maximal value 1. Let’s use some examples to analyze the GMS induced LQM. The most commonly encountered distortions in many real image processing systems are JPEG compression, JPEG2000 compression, additive white noise (AWN) and Gaussian blur (GB). In Fig. 2, for each of the four types of distortions, two reference images with different contents and their corresponding distorted images are shown (the images are selected from the LIVE database [11]). Their gradient magnitude images ($\mathbf{m_r}$ and $\mathbf{m_d}$) and the corresponding GMS maps are also shown. In the GMS map, the brighter the gray level, the higher the similarity, and thus the higher the predicted local quality. These images contain a variety of important structures such as large scale edges, smooth areas and fine textures, etc. A good IQA model should be adaptable to the broad array of possible natural scenes and local structures.

In Fig. 2, examples of structure degradation are shown in the gradient magnitude domain. Typical areas are highlighted with red curves. From the first group, it can be seen that the artifacts caused by AWN are masked in the large structure and texture areas, while the artifacts are more visible in flat areas. This is broadly consistent with human perception. In the second group, the degradations caused by JPEG compression are mainly blocking effects (see the background area of image *parrots* and the wall area of image *house*) and loss of fine details. Clearly, the GMS map is highly responsive to these distortions. Regarding JPEG2000 compression, artifacts are introduced in the vicinity of edge structures and in the textured areas. Regarding GB, the whole GMS map is clearly changed after image distortion. All these observations imply that the image gradient magnitude is a highly relevant feature for the task of IQA.

### *B. Pooling with Standard Deviation*

The LQM reflects the local quality of each small patch in the distorted image. The image overall quality score can then be estimated from the LQM via a pooling stage. The most commonly used pooling strategy is average pooling, i.e., simply averaging the LQM values as the final IQA score. We refer to the IQA model by applying average pooling to the GMS map as Gradient Magnitude Similarity Mean (GMSM):

$$GMSM = \frac{1}{N}\sum_{i=1}^{N} GMS(i) \quad (5)$$

where $N$ is the total number of pixels in the image. Clearly, a higher GMSM score means higher image quality. Average pooling assumes that each pixel has the same importance in estimating the overall image quality. As introduced in Section I, researchers have devoted much effort to design weighted pooling methods [9, 10, 16, 19, 20 and 29]; however, the improvement brought by weighted pooling over average pooling is not always significant [31] and the computation of weights can be costly.

We propose a new pooling strategy with the GMS map. A natural image generally has a variety of local structures in its scene. When an image is distorted, the different local structures will suffer different degradations in gradient magnitude. This is an inherent property of natural images. For example, the distortions introduced by JPEG2000 compression include blocking, ringing, blurring, etc. Blurring will cause less quality degradation in flat areas than in textured areas, while blocking will cause higher quality degradation in flat areas than in textured areas. However, the average pooling strategy ignores this fact and it cannot reflect how the local quality degradation varies. Based on the idea that the global variation of image local quality degradation can reflect its overall quality, we propose to compute the standard deviation of the GMS map and take it as the final IQA index, namely Gradient Magnitude Similarity Deviation (GMSD):

$$GMSD = \sqrt{\frac{1}{N}\sum_{i=1}^{N}\left(GMS(i) - GMSM\right)^2} \quad (6)$$

Note that the value of GMSD reflects the range of distortion severities in an image. The higher the GMSD score, the larger the distortion range, and thus the lower the image perceptual quality.

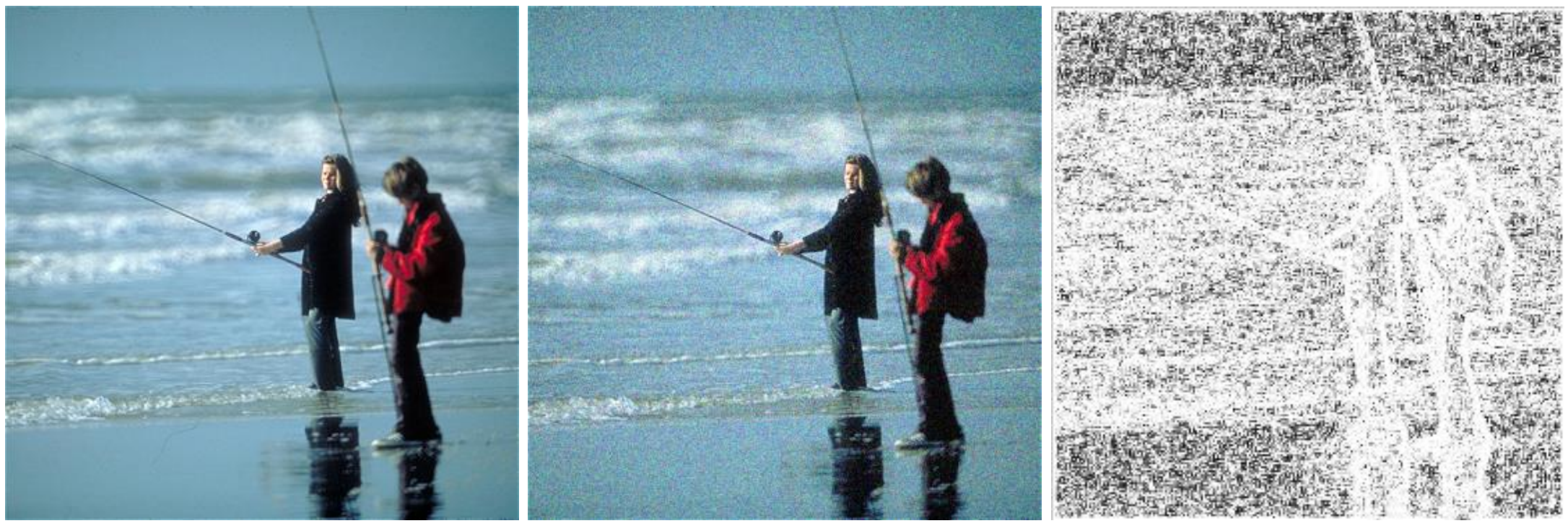

(a)

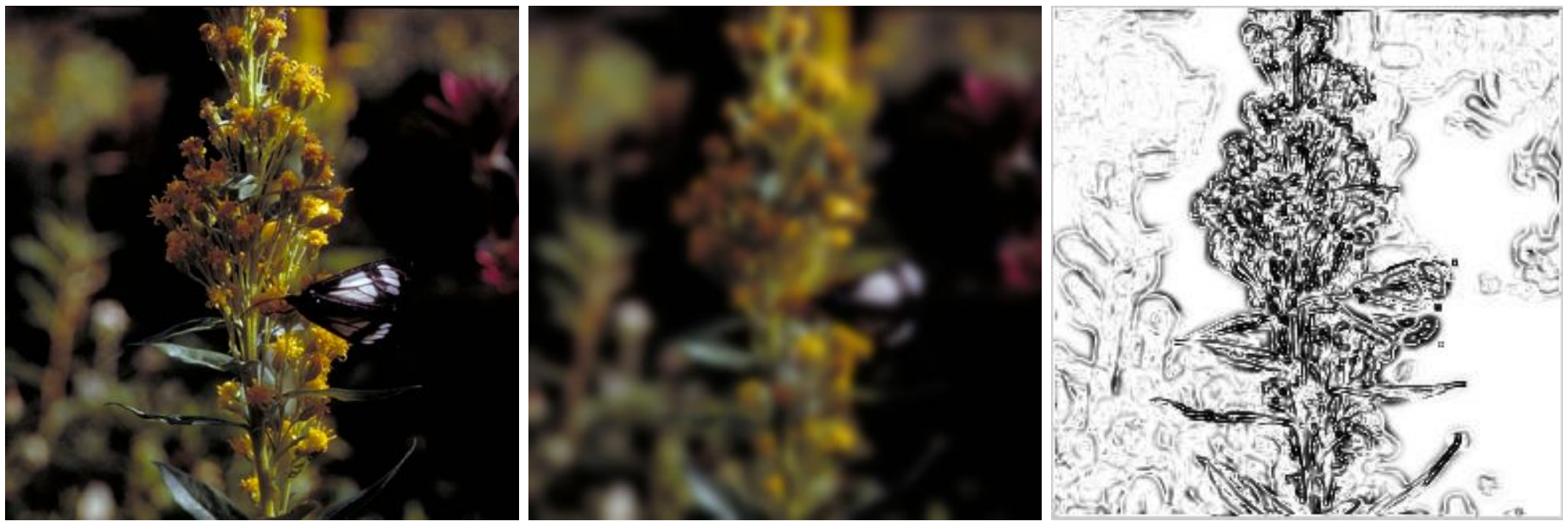

(b)

**Figure 3** Comparison beween GMSM and GMSD as a subjective quality indicator. Note that like DMOS, GMSD is a distortion index (a lower DMOS/GMSD means higher quality), while GMSM is a quality index (a highr GMSM means higher quality). (a) Original image *Fishing*, its Gaussian noise contaminated version (DMOS=0.4403; GMSM=0.8853; GMSD=0.1420) and their gradient simiarity map. (b) Original image *Flower*, its blurred version (DMOS=0.7785; GMSM=0.8745; GMSD=0.1946) and their gradient simiarity map. Based on the human subjective DMOS, image *Fishing* has much higher quality than image *Flower*. GMSD gives the correct judgement but GMSM fails.

In Fig. 3, we show two reference images from the CSIQ database [12], their distorted images and the corresponding GMS maps. The first image *Fishing* is corrupted by additive white noise, and the second image *Flower* is Gaussian blurred. From the GMS map of distorted image *Fishing*, one can see that its local quality is more homogenous, while from the GMS map of distorted image *Flower*, one can see that its local quality in the center area is much worse than at other areas. The human subjective DMOS scores of the two distorted images are 0.4403 and 0.7785, respectively, indicating that the quality of the first image is obviously better than the second one. (Note that like GMSD, DMOS also measures distortion; the lower it is, the better the image quality.) By using GMSM, however, the predicted quality scores of the two images are 0.8853 and 0.8745, respectively, indicating that the perceptual quality of the first image is similar to the second one, which is inconsistent with the subjective DMOS scores.

By using GMSD, the predicted quality scores of the two images are 0.1420 and 0.1946, respectively, which is a consistent judgment relative to the subjective DMOS scores, i.e., the first distorted image has better quality than the second one. More examples of the consistency between GMSD and DMOS can be found in Fig. 2. For each distortion type, the two images of different contents have similar DMOS scores, while their GMSD indices are also very close. These examples validate that the deviation pooling strategy coupled with the GMS quality map can accurately predict the perceptual image quality.

## III. Experiments and Results Analysis

### A. Databases and Evaluation Protocols

The performance of an IQA model is typically evaluated from three aspects regarding its prediction power [21]: prediction *accuracy*, prediction *monotonicity*, and prediction *consistency*. The computation of these indices requires a regression procedure to reduce the nonlinearity of predicted scores. We denote by $Q$, $Q_p$ and $S$ the vectors of the original IQA scores, the IQA scores after regression and the subjective scores, respectively. The logistic regression function is employed for the nonlinear regression [21]:

$$Q_p = \beta_1\left(\frac{1}{2} - \frac{1}{\exp(\beta_2(Q-\beta_3))}\right) + \beta_4 Q + \beta_5 \quad (7)$$

where $\beta_1$, $\beta_2$, $\beta_3$, $\beta_4$ and $\beta_5$ are regression model parameters.

After the regression, 3 correspondence indices can be computed for performance evaluation [21]. The first one is the Pearson linear Correlation Coefficient (PCC) between $Q_p$ and $S$, which is to evaluate the prediction accuracy:

$$PCC(Q_P, S) = \frac{\bar{Q}_P^T \bar{S}}{\sqrt{\bar{Q}_P^T \bar{Q}_P \bar{S}^T \bar{S}}} \quad (8)$$

where $\bar{Q}_P$ and $\bar{S}$ are the mean-removed vectors of $Q_P$ and $S$, respectively, and subscript "$T$" means transpose. The second index is the Spearman Rank order Correlation coefficient (SRC) between $Q$ and $S$, which is to evaluate the prediction monotonicity:

$$SRC(Q, S) = 1 - \frac{6\sum_{i=1}^{n} d_i^2}{n(n^2 - 1)} \quad (9)$$

where $d_i$ is the difference between the ranks of each pair of samples in $Q$ and $S$, and $n$ is the total number of samples. Note that the logistic regression does not affect the SRC index, and we can compute it before regression. The third index is the root mean square error (RMSE) between $Q_p$ and $S$, which is to evaluate the prediction consistency:

$$RMSE(Q_P, S) = \sqrt{(Q_P - S)^T (Q_P - S) / n} \quad (10).$$

With the SRC, PCC and RMSE indices, we evaluate the IQA models on three large scale and publicly accessible IQA databases: LIVE [11], CSIQ [12], and TID2008 [13]. The LIVE database consists of 779 distorted images generated from 29 reference images. Five types of distortions are applied to the reference images at various levels: JPEG2000 compression, JPEG compression, additive white noise (AWN), Gaussian blur (GB) and simulated fast fading Rayleigh channel (FF). These distortions reflect a broad range of image impairments, for example, edge smoothing, block artifacts and random noise. The CSIQ database consists of 30 reference images and their distorted counterparts with six types of distortions at five different distortion levels. The six types of distortions include JPEG2000, JPEG, AWN, GB, global contrast decrements (CTD), and additive pink Gaussian noise (PGN). There are a total of 886 distorted images in it. The TID2008 database is the largest IQA database to date. It has 1,700 distorted images, generated from 25 reference images with 17 types of distortions at 4 levels. Please refer to [13] for details of the distortions. Each image in these databases has been evaluated by human subjects under controlled conditions, and then assigned a quantitative subjective quality score: Mean Opinion Score (MOS) or Difference MOS (DMOS).

To demonstrate the performance of GMSD, we compare it with 11 state-of-the-art and representative FR-IQA models, including PSNR, IFC [22], VIF [23], SSIM [8], MS-SSIM [17], MAD [12], FSIM [7], IW-SSIM [16], G-SSIM [6], GSD [5] and GS [15]. Among them, FSIM, G-SSIM, GSD and GS explicitly exploit gradient information. Except for G-SSIM and GSD, which are implemented by us, the source codes of all the other models were obtained from the original authors. To more clearly demonstrate the effectiveness of the proposed deviation pooling strategy, we also present the results of GMSM which uses average pooling. As in most of the previous literature [7-8, 16-17], all of the competing algorithms are applied to the luminance channel of the test images.

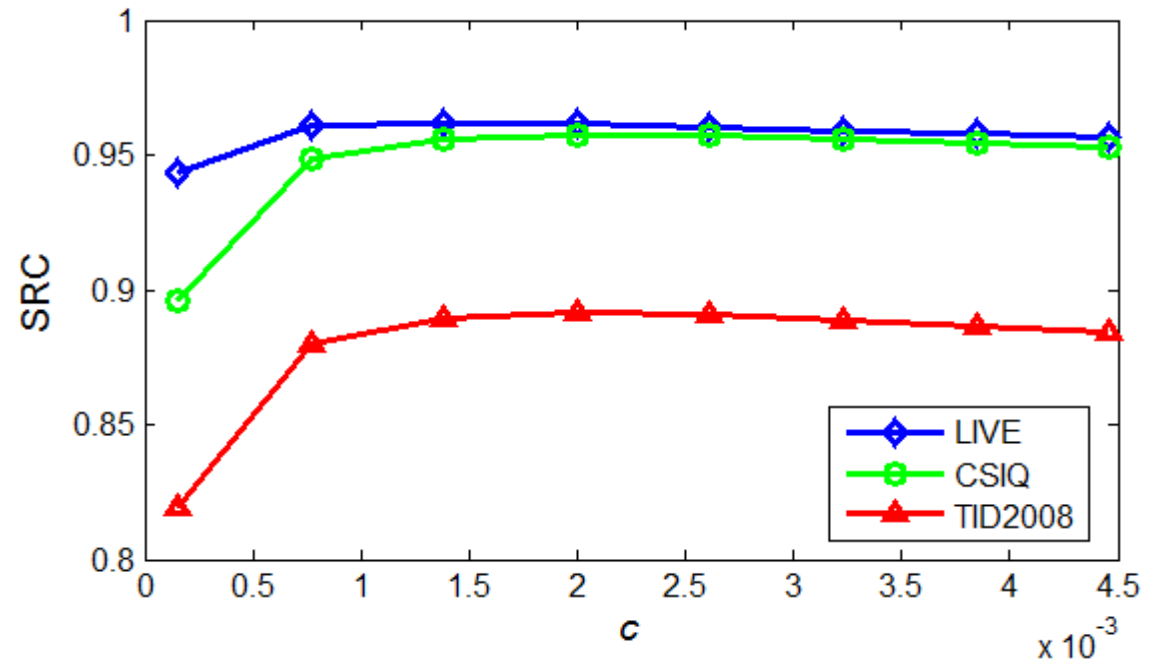

**Figure 4** The performance of GMSD in terms of SRC vs. constant $k$ on the three databases.

### B. *Implementation of GMSD*

The only parameter in the proposed GMSM and GMSD models is the constant c in Eq. (4). Apart from ensuring the numerical stability, the constant c also plays a role in mediating the contrast response in low gradient areas. We normalize the pixel values of 8-bit luminance image into range [0, 1]. Fig. 4 plots the SRC curves against c by applying GMSD to the LIVE, CSIQ and TID2008 databases. One can see that for all the databases, GMSD shows similar preference to the value of c. In our implementation, we set c=0.0026. In addition, as in the implementations of SSIM [8] and FSIM [7], the images r and d are first filtered by a 2×2 average filter, and then down-sampled by a factor of 2. MATLAB source code that implements GMSD can be downloaded at http://www4.comp.polyu.edu.hk /~cslzhang/IQA/GMSD/GMSD.htm.

### C. *Performance Comparison*

In Table I, we compare the competing IQA models' performance on each of the three IQA databases in terms of SRC, PCC and RMSE. The top three models for each evaluation criterion are shown in boldface. We can see that the top models are mostly GMSD (9 times), FSIM (7 times), IW-SSIM (6 times) and VIF (5 times). In terms of all the three criteria (SRC, PCC and RMSE), the proposed GMSD outperforms all the other models on the TID2008 and CSIQ databases. On the LIVE database, VIF, FSIM and GMSD perform almost the same. Compared with gradient based models such as GSD, G-SSIM and GS, GMSD outperforms them by a large margin. Compared with GMSM, the superiority of GMSD is obvious, demonstrating that the proposed deviation pooling strategy works much better than the average pooling strategy on the GMS induced LQM. The FSIM algorithm also employs gradient similarity. It has similar results to GMSD on the LIVE and TID2008 databases, but lags GMSD on the CSIQ database with a lower SRC/PCC and larger RMSE.

In Fig. 5, we show the scatter plots of predicted quality scores against subjective DMOS scores for some representative models (PSNR, VIF, GS, IW-SSIM, MS-SSIM, MAD, FSIM,

**Table I:** Performance of the proposed GMSD and the other eleven competing FR-IQA models in terms of SRC, PCC, and RMSE on the 3 databases. The top three models for each criterion are shown in boldface.

| *IQA model* | LIVE (779 images) | | | CSIQ (886 images) | | | TID2008 (1700 images) | | | Weighted Average | |
|---|---|---|---|---|---|---|---|---|---|---|---|
| | SRC | PCC | RMSE | SRC | PCC | RMSE | SRC | PCC | RMSE | SRC | PCC |
| *PSNR* | 0.876 | 0.872 | 13.36 | 0.806 | 0.751 | 0.173 | 0.553 | 0.523 | 1.144 | 0.694 | 0.664 |
| *IFC [22]* | 0.926 | 0.927 | 10.26 | 0.767 | 0.837 | 0.144 | 0.568 | 0.203 | 1.314 | 0.703 | 0.537 |
| *GSD [5]* | 0.908 | 0.913 | 11.149 | 0.854 | 0.854 | 0.137 | 0.657 | 0.707 | 0.949 | 0.766 | 0.793 |
| *G-SSIM [6]* | 0.918 | 0.920 | 10.74 | 0.872 | 0.874 | 0.127 | 0.731 | 0.760 | 0.873 | 0.811 | 0.827 |
| *SSIM [8]* | 0.948 | 0.945 | 8.95 | 0.876 | 0.861 | 0.133 | 0.775 | 0.773 | 0.851 | 0.841 | 0.836 |
| *VIF [23]* | **0.964** | **0.960** | **7.61** | 0.919 | **0.928** | **0.098** | 0.749 | 0.808 | 0.790 | 0.844 | 0.875 |
| *MAD [12]* | 0.944 | 0.939 | 9.37 | 0.899 | 0.820 | 0.150 | 0.771 | 0.748 | 0.891 | 0.845 | 0.811 |
| *MS-SSIM [17]* | 0.952 | 0.950 | 8.56 | 0.877 | 0.659 | 0.197 | 0.809 | 0.801 | 0.803 | 0.860 | 0.798 |
| *GS [15]* | 0.956 | 0.951 | 8.43 | 0.911 | 0.896 | 0.116 | 0.850 | 0.842 | 0.723 | 0.891 | 0.882 |
| *GMSM* | **0.960** | 0.956 | 8.049 | 0.929 | 0.913 | 0.107 | 0.848 | 0.837 | 0.735 | 0.895 | 0.884 |
| *IW-SSIM [16]* | 0.957 | 0.952 | 8.35 | **0.921** | **0.914** | **0.106** | **0.856** | **0.858** | **0.689** | **0.896** | **0.895** |
| *FSIM [7]* | **0.963** | **0.960** | **7.67** | **0.924** | 0.912 | 0.108 | **0.880** | **0.874** | **0.653** | **0.911** | **0.904** |
| ***GMSD*** | **0.960** | **0.960** | **7.62** | **0.957** | **0.954** | **0.079** | **0.891** | **0.879** | **0.640** | **0.924** | **0.917** |

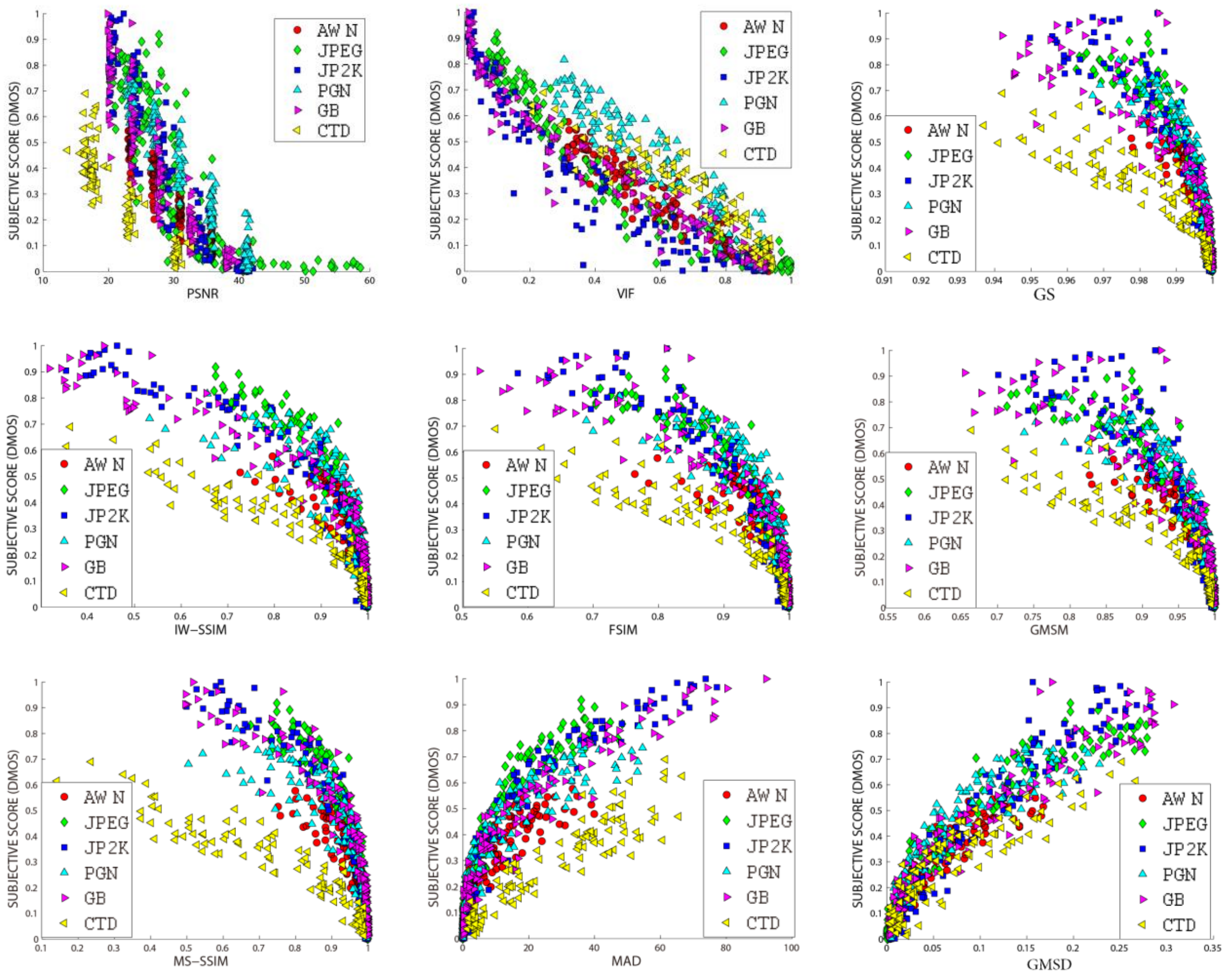

**Figure 5** Scatter plots of predicted quality scores against the subjective quality scores (DMOS) by representative FR-IQA models on the CSIQ database. The six types of distortions are represented by different shaped colors.

| LIVE | PSNR | IFC | GSD | G-SSIM | SSIM | VIF | MAD | MS-SSIM | GS | GMSM | IW-SSIM | FSIM | **GMSD** |
|---|---|---|---|---|---|---|---|---|---|---|---|---|---|
| *PSNR* | 0 | 0 | 0 | 0 | 0 | 0 | 0 | 0 | 0 | 0 | 0 | 0 | 0 |
| *IFC* | 1 | 0 | 1 | 0 | 0 | 0 | 0 | 0 | 0 | 0 | 0 | 0 | 0 |
| *GSD* | 1 | 0 | 0 | 0 | 0 | 0 | 0 | 0 | 0 | 0 | 0 | 0 | 0 |
| *G-SSIM* | 1 | 0 | 0 | 0 | 0 | 0 | 0 | 0 | 0 | 0 | 0 | 0 | 0 |
| *SSIM* | 1 | 1 | 1 | 1 | 0 | 0 | 0 | 0 | 0 | 0 | 0 | 0 | 0 |
| *VIF* | 1 | 1 | 1 | 1 | 1 | 0 | 1 | 1 | 1 | 0 | 1 | 0 | 0 |
| *MAD* | 1 | 1 | 1 | 1 | 0 | 0 | 0 | 0 | 0 | 0 | 0 | 0 | 0 |
| *MS-SSIM* | 1 | 1 | 1 | 1 | 0 | 0 | 1 | 0 | 0 | 0 | 0 | 0 | 0 |
| *GS* | 1 | 1 | 1 | 1 | 1 | 0 | 1 | 0 | 0 | 0 | 0 | 0 | 0 |
| *GMSM* | 1 | 1 | 1 | 1 | 1 | 0 | 1 | 1 | 0 | 0 | 0 | 0 | 0 |
| *IW-SSIM* | 1 | 1 | 1 | 1 | 1 | 0 | 1 | 0 | 0 | 0 | 0 | 0 | 0 |
| *FSIM* | 1 | 1 | 1 | 1 | 1 | 0 | 1 | 1 | 1 | 0 | 1 | 0 | 0 |
| ***GMSD*** | 1 | 1 | 1 | 1 | 1 | 0 | 1 | 1 | 1 | 0 | 1 | 0 | 0 |

(a)

| CSIQ | PSNR | IFC | GSD | G-SSIM | SSIM | VIF | MAD | MS-SSIM | GS | GMSM | IW-SSIM | FSIM | **GMSD** |
|---|---|---|---|---|---|---|---|---|---|---|---|---|---|
| *PSNR* | 0 | 0 | 0 | 0 | 0 | 0 | 0 | 1 | 0 | 0 | 0 | 0 | 0 |
| *IFC* | 1 | 0 | 0 | 0 | 0 | 0 | 0 | 1 | 0 | 0 | 0 | 0 | 0 |
| *GSD* | 1 | 0 | 0 | 0 | 0 | 0 | 1 | 1 | 0 | 0 | 0 | 0 | 0 |
| *G-SSIM* | 1 | 1 | 1 | 0 | 0 | 0 | 1 | 1 | 0 | 0 | 0 | 0 | 0 |
| *SSIM* | 1 | 1 | 0 | 0 | 0 | 0 | 1 | 1 | 0 | 0 | 0 | 0 | 0 |
| *VIF* | 1 | 1 | 1 | 1 | 1 | 0 | 1 | 1 | 1 | 1 | 1 | 1 | 0 |
| *MAD* | 1 | 0 | 0 | 0 | 0 | 0 | 0 | 1 | 0 | 0 | 0 | 0 | 0 |
| *MS-SSIM* | 0 | 0 | 0 | 0 | 0 | 0 | 0 | 0 | 0 | 0 | 0 | 0 | 0 |
| *GS* | 1 | 1 | 1 | 1 | 1 | 0 | 1 | 1 | 0 | 0 | 0 | 0 | 0 |
| *GMSM* | 1 | 1 | 1 | 1 | 1 | 0 | 1 | 1 | 1 | 0 | 0 | 0 | 0 |
| *IW-SSIM* | 1 | 1 | 1 | 1 | 1 | 0 | 1 | 1 | 1 | 0 | 0 | 0 | 0 |
| *FSIM* | 1 | 1 | 1 | 1 | 1 | 0 | 1 | 1 | 1 | 0 | 0 | 0 | 0 |
| ***GMSD*** | 1 | 1 | 1 | 1 | 1 | 1 | 1 | 1 | 1 | 1 | 1 | 1 | 0 |

(b)

| TID2008 | PSNR | IFC | GSD | G-SSIM | SSIM | VIF | MAD | MS-SSIM | GS | GMSM | IW-SSIM | FSIM | **GMSD** |
|---|---|---|---|---|---|---|---|---|---|---|---|---|---|
| *PSNR* | 0 | 1 | 0 | 0 | 0 | 0 | 0 | 0 | 0 | 0 | 0 | 0 | 0 |
| *IFC* | 0 | 0 | 0 | 0 | 0 | 0 | 0 | 0 | 0 | 0 | 0 | 0 | 0 |
| *GSD* | 1 | 1 | 0 | 0 | 0 | 0 | 0 | 0 | 0 | 0 | 0 | 0 | 0 |
| *G-SSIM* | 1 | 1 | 1 | 0 | 0 | 0 | 0 | 0 | 0 | 0 | 0 | 0 | 0 |
| *SSIM* | 1 | 1 | 1 | 0 | 0 | 0 | 1 | 0 | 0 | 0 | 0 | 0 | 0 |
| *VIF* | 1 | 1 | 1 | 1 | 1 | 0 | 1 | 0 | 0 | 0 | 0 | 0 | 0 |
| *MAD* | 1 | 1 | 1 | 0 | 0 | 0 | 0 | 0 | 0 | 0 | 0 | 0 | 0 |
| *MS-SSIM* | 1 | 1 | 1 | 1 | 1 | 0 | 1 | 0 | 0 | 0 | 0 | 0 | 0 |
| *GS* | 1 | 1 | 1 | 1 | 1 | 1 | 1 | 1 | 0 | 0 | 0 | 0 | 0 |
| *GMSM* | 1 | 1 | 1 | 1 | 1 | 1 | 1 | 1 | 0 | 0 | 0 | 0 | 0 |
| *IW-SSIM* | 1 | 1 | 1 | 1 | 1 | 1 | 1 | 1 | 1 | 1 | 0 | 0 | 0 |
| *FSIM* | 1 | 1 | 1 | 1 | 1 | 1 | 1 | 1 | 1 | 1 | 1 | 0 | 0 |
| ***GMSD*** | 1 | 1 | 1 | 1 | 1 | 1 | 1 | 1 | 1 | 1 | 1 | 0 | 0 |

(c)

**Figure 6** The results of statistical significance tests of the competing IQA models on the (a) LIVE, (b) CSIQ and (c) TID2008 databases. A value of '1' (highlighted in green) indicates that the model in the row is significantly better than the model in the column, while a value of '0' (highlighted in red) indicates that the first model is not significantly better than the second one. Note that the proposed GMSD is significantly better than most of the competitors on all the three databases, while no IQA model is significantly better than GMSD.

**Table II:** Performance comparison of the IQA models on each individual distortion type in terms of SRC.

| | | *PSNR* | *IFC* | *GSD* | *G-SSIM* | *SSIM* | *VIF* | *MAD* | *MS-SSIM* | *GS* | *GMSM* | *IW-SSIM* | *FSIM* | *GMSD* |
|---|---|---|---|---|---|---|---|---|---|---|---|---|---|---|
| *LIVE database* | *JP2K* | 0.895 | 0.911 | 0.911 | 0.935 | 0.961 | **0.970** | 0.964 | 0.963 | **0.970** | 0.968 | 0.965 | **0.971** | **0.971** |
| | *JPEG* | 0.881 | 0.947 | 0.931 | 0.944 | 0.976 | **0.985** | 0.975 | 0.982 | 0.978 | 0.979 | **0.981** | **0.983** | 0.978 |
| | *AWN* | **0.985** | 0.938 | 0.879 | 0.926 | 0.969 | **0.986** | **0.986** | 0.977 | 0.977 | 0.967 | 0.967 | 0.965 | 0.974 |
| | *GB* | 0.782 | 0.958 | 0.964 | 0.968 | 0.952 | **0.973** | 0.933 | 0.955 | 0.952 | 0.959 | **0.972** | **0.971** | 0.957 |
| | *FF* | 0.891 | **0.963** | 0.953 | 0.948 | **0.956** | **0.965** | **0.956** | 0.941 | 0.940 | 0.943 | 0.944 | 0.950 | 0.942 |
| *CSIQ database* | *AWN* | 0.936 | 0.843 | 0.732 | 0.810 | 0.897 | 0.957 | **0.960** | 0.944 | 0.944 | 0.962 | 0.938 | 0.926 | **0.968** |
| | *JPEG* | 0.888 | 0.941 | 0.927 | 0.927 | 0.954 | 0.970 | **0.967** | 0.964 | 0.963 | 0.959 | **0.966** | **0.966** | 0.965 |
| | *JP2K* | 0.936 | 0.925 | 0.913 | 0.932 | 0.960 | 0.967 | **0.977** | **0.972** | 0.965 | 0.957 | 0.968 | 0.968 | **0.972** |
| | *PGN* | 0.934 | 0.826 | 0.731 | 0.796 | 0.892 | **0.951** | **0.954** | 0.933 | 0.939 | 0.945 | 0.906 | 0.923 | **0.950** |
| | *GB* | 0.929 | 0.953 | 0.960 | 0.958 | 0.961 | **0.974** | 0.966 | **0.975** | 0.959 | 0.958 | **0.978** | 0.972 | 0.971 |
| | *CTD* | 0.862 | 0.487 | **0.948** | 0.851 | 0.793 | 0.934 | 0.917 | **0.945** | 0.936 | 0.933 | **0.954** | 0.942 | 0.904 |
| *TID2008 database* | *AWN* | **0.907** | 0.581 | 0.535 | 0.574 | 0.811 | 0.880 | 0.864 | 0.812 | 0.861 | **0.887** | 0.787 | 0.857 | **0.918** |
| | *ANMC* | **0.899** | 0.546 | 0.479 | 0.556 | 0.803 | 0.876 | 0.839 | 0.807 | 0.809 | **0.877** | 0.792 | 0.851 | **0.898** |
| | *SCN* | **0.917** | 0.596 | 0.568 | 0.600 | 0.815 | 0.870 | 0.898 | 0.826 | 0.894 | 0.877 | 0.771 | 0.848 | **0.913** |
| | *MN* | **0.852** | 0.673 | 0.586 | 0.609 | 0.779 | **0.868** | 0.734 | 0.802 | 0.745 | 0.760 | **0.809** | 0.802 | 0.709 |
| | *HFN* | **0.927** | 0.732 | 0.661 | 0.728 | 0.873 | 0.907 | 0.896 | 0.871 | 0.895 | **0.915** | 0.866 | 0.909 | **0.919** |
| | *IMN* | **0.872** | 0.534 | 0.577 | 0.409 | 0.673 | **0.833** | 0.513 | 0.698 | 0.723 | **0.748** | 0.646 | 0.746 | 0.661 |
| | *QN* | **0.870** | 0.586 | 0.609 | 0.672 | 0.853 | 0.797 | 0.850 | 0.852 | **0.880** | 0.867 | 0.818 | 0.855 | **0.887** |
| | *GB* | 0.870 | 0.856 | 0.911 | 0.924 | **0.954** | **0.954** | 0.914 | **0.954** | **0.960** | 0.952 | **0.964** | 0.947 | 0.897 |
| | *DEN* | 0.942 | 0.797 | 0.878 | 0.880 | 0.953 | 0.916 | 0.945 | 0.961 | **0.972** | **0.966** | 0.947 | 0.960 | **0.975** |
| | *JPEG* | 0.872 | 0.818 | 0.839 | 0.859 | 0.925 | 0.917 | **0.942** | **0.939** | **0.939** | **0.939** | 0.918 | 0.928 | **0.952** |
| | *JP2K* | 0.813 | 0.944 | 0.923 | 0.944 | 0.962 | 0.971 | 0.972 | 0.970 | **0.976** | 0.973 | 0.974 | **0.977** | **0.980** |
| | *JGTE* | 0.752 | 0.791 | **0.880** | 0.855 | 0.868 | 0.859 | 0.851 | 0.872 | **0.879** | **0.882** | 0.859 | 0.871 | 0.862 |
| | *J2TE* | 0.831 | 0.730 | 0.722 | 0.758 | 0.858 | 0.850 | 0.840 | 0.861 | **0.894** | **0.877** | 0.820 | 0.854 | **0.883** |
| | *NEPN* | 0.581 | **0.842** | 0.770 | 0.754 | 0.711 | 0.762 | **0.837** | 0.752 | 0.739 | 0.744 | 0.772 | 0.749 | 0.760 |
| | *Block* | 0.619 | 0.677 | 0.811 | 0.810 | 0.846 | 0.832 | 0.159 | 0.499 | **0.886** | **0.899** | 0.762 | 0.849 | **0.897** |
| | *MS* | 0.696 | 0.425 | 0.441 | 0.715 | **0.723** | 0.510 | 0.587 | **0.773** | **0.719** | 0.630 | 0.707 | 0.669 | 0.649 |
| | *CTC* | 0.586 | 0.171 | 0.573 | 0.552 | 0.525 | **0.819** | 0.493 | 0.625 | **0.669** | **0.663** | 0.630 | 0.648 | 0.466 |

GMSM and GMSD) on the CSIQ database, which has six types of distortions (AWN, JPEG, JPEG2000, PGN, GB and CTD). One can observe that for FSIM, MAD, MS-SSIM, GMSM, IW-SSIM and GS, the distribution of predicted scores on the CTD distortion deviates much from the distributions on other types of distortions, degrading their overall performance. When the distortion is severe (i.e., large DMOS values), GS, GMSM and PSNR yield less accurate quality predictions. The information fidelity based VIF performs very well on the LIVE database but not very well on the CSIQ and TID2008 databases. This is mainly because VIF does not predict the images' quality consistently across different distortion types on these two databases, as can be observed from the scatter plots with CSIQ database in Fig. 5.

In Table I, we also show the weighted average of SRC and PCC scores by the competing FR-IQA models over the three databases, where the weights were determined by the sizes (i.e., number of images) of the three databases. According to this, the top 3 models are GMSD, FSIM and IW-SSIM. Overall, the proposed GMSD achieves outstanding and consistent performance across the three databases.

In order to make statistically meaningful conclusions on the models' performance, we further conducted a series of hypothesis tests based on the prediction residuals of each model after nonlinear regression. The results of significance tests are shown in Fig. 6. By assuming that the model's prediction residuals follow the Gaussian distribution (the Jarque-Bera test [35] shows that only 3 models on LIVE and 4 models on CSIQ violate this assumption), we apply the left-tailed $F$-test to the residuals of every two models to be compared. A value of $H$=1 for the left-tailed $F$-test at a significance level of 0.05 means that the first model (indicated by the row in Fig. 6) has better IQA performance than the second model (indicated by the column in Fig. 6) with a confidence greater than 95%. A value of $H$=0 means that the first model is not significantly better than the second one. If $H$=0 always holds no matter which one of the two models is taken as the first one, then the two models have no significant difference in performance. Figs. 6(a) ~ 6(c) show the significance test results on the LIVE, CSIQ and TID2008 databases, respectively. We see that on the LIVE database, GMSD is significantly better than all the other IQA models except for VIF, GMSM and FSIM. On the CSIQ database, GMSD is significantly better than all the other models. On the TID2008 database, GMSD is significantly better than all the other IQA models except for FSIM. Note that on all the three databases, no IQA model performs significantly better than GMSD.

### *D. Performance Comparison on Individual Distortion Types*

To more comprehensively evaluate an IQA model's ability to predict image quality degradations caused by specific types of distortions, we compare the performance of competing methods on each type of distortion. The results are listed in Table II. To save space, only the SRC scores are shown. There are a total of 28 groups of distorted images in the three databases. In Table II, we use boldface font to highlight the top 3 models in each group. One can see that GMSD is among the top 3 models 14 times, followed by GS and VIF, which are among the top 3 models 11 and 10 times, respectively. However, neither GS nor VIF ranks among the top 3 in terms of overall performance on the 3 databases. The classical PSNR also performs among the top 3 for 8 groups, and a common point of these 8 groups is that they are all noise contaminated. PSNR is, indeed, an effective measure of perceptual quality of noisy images. However, PSNR is not able to faithfully measure the quality of images impaired by other types of distortions.

Generally speaking, performing well on specific types of distortions does not guarantee that an IQA model will perform well on the whole database with a broad spectrum of distortion types. A good IQA model should also predict the image quality consistently across different types of distortions. Referring to the scatter plots in Fig. 5, it can be seen that the scatter plot of GMSD is more concentrated across different groups of distortion types. For example, its points corresponding to JPEG2000 and PGN distortions are very close to each other. However, the points corresponding to JPEG2000 and PGN for VIF are relatively far from each other. We can have similar observations for GS on the distortion types of PGN and CTD. This explains why some IQA models perform well for many individual types of distortions but they do not perform well on the entire databases; that is, these IQA models behave rather differently on different types of distortions, which can be attributed to the different ranges of quality scores for those distortion types [43].

The gradient based models G-SSIM and GSD do not show good performance on either many individual types of distortions or the entire databases. G-SSIM computes the local variance and covariance of gradient magnitude to gauge contrast and structure similarities. This may not be an effective use of gradient information. The gradient magnitude describes the local contrast of image intensity; however, the image local structures with different distortions may have similar variance of gradient magnitude, making G-SSIM less effective to distinguish those distortions. GSD combines the orientation differences of gradient, the contrast similarity and the gradient similarity; however, there is intersection between these kinds of information, making GSD less discriminative of image quality. GMSD only uses the gradient magnitude information but achieves highly competitive results against the competing methods. This validates that gradient magnitude, coupled with the deviation pooling strategy, can serve as an excellent predictive image quality feature.

**Table III:** SRC results of SD pooling on some representative IQA models.

| *Database* | (Weighted) average pooling | | | SD pooling | | | Performance gain | | |
|---|---|---|---|---|---|---|---|---|---|
| | LIVE | CSIQ | TID2008 | LIVE | CSIQ | TID2008 | LIVE | CSIQ | TID2008 |
| *MSE* | 0.876 | 0.806 | 0.553 | 0.877 | 0.834 | 0.580 | 0.18% | 3.55% | 4.88% |
| *SSIM [8]* | 0.948 | 0.876 | 0.775 | 0.917 | 0.817 | 0.756 | -3.22% | -6.71% | -2.44% |
| *MS-SSIM [17]* | 0.952 | 0.877 | 0.809 | 0.921 | 0.826 | 0.650 | -3.28% | -5.86% | -19.71% |
| *FSIM [7]* | 0.963 | 0.924 | 0.880 | 0.960 | 0.956 | 0.892 | -0.33% | 3.52% | 1.26% |
| *G-SSIM [6]* | 0.918 | 0.872 | 0.731 | 0.763 | 0.757 | 0.708 | -16.93% | -13.20% | -3.09% |
| *GSD [5]* | 0.914 | 0.828 | 0.576 | 0.669 | 0.611 | 0.568 | -26.76% | -26.20% | -1.36% |

### E. Standard Deviation Pooling on Other IQA models

As shown in previous sections, the method of standard deviation (SD) pooling applied to the GMS map leads to significantly elevated performance of image quality prediction. It is therefore natural to wonder whether the SD pooling strategy can deliver similar performance improvement on other IQA models. To explore this, we modified six representative FR-IQA methods, all of which are able to generate an LQM of the test image: MSE (which is equivalent to PSNR but can produce an LQM), SSIM [8], MS-SSIM [17], FSIM [7], G-SSIM [6] and GSD [5]. The original pooling strategies of these methods are either average pooling or weighted pooling. For MSE, SSIM, G-SSIM, GSD and FSIM, we directly applied the SD pooling to their LQMs to yield the predicted quality scores. For MS-SSIM, we applied SD pooling to its LQM on each scale, and then computed the product of the predicted scores on all scales as the final score. In Table III, the SRC results of these methods by using their nominal pooling strategies and the SD pooling strategy are listed.

Table III makes it clear that except for MSE, all the other IQA methods fail to gain in performance by using SD pooling instead of their nominal pooling strategies. The reason may be that in these methods, the LQM is generated using multiple, diverse types of features. The interaction between these features may complicate the estimation of image local quality so that SD pooling does not apply. By contrast, MSE and GMSD use only the original intensity and the intensity of gradient magnitude, respectively, to calculate the LQM.

### F. Complexity

In applications such as real-time image/video quality monitoring and prediction, the complexity of implemented IQA models becomes crucial. We thus analyze the computational complexity of GMSD, and then compare the competing IQA models in terms of running time.

Suppose that an image has $N$ pixels. The classical PSNR has the lowest complexity, and it only requires $N$ multiplications and $2N$ additions. The main operations in the proposed GMSD model include calculating image gradients (by convolving the image with two 3×3 template integer filters), thereby producing gradient magnitude maps, generating the GMS map, and deviation pooling. Overall, it requires $19N$ multiplications and $16N$ additions to yield the final quality score. Meanwhile, it only needs to store at most 4 directional gradient images (each of size $N$) in memory (at the gradient calculation stage). Therefore, both the time and memory complexities of GMSD are $O(N)$. In other words, the time and memory cost of GMSD scales linearly with image size. This is a very attractive property since image resolutions have been rapidly increasing with the development of digital imaging technologies. In addition, the computation of image gradients and GMS map can be parallelized by partitioning the reference and distorted images into blocks if the image size is very large.

**Table IV:** Running time of the competing IQA models.

| *Models* | Running time (s) |
|---|---|
| *MAD [12]* | 2.0715 |
| *IFC [22]* | 1.1811 |
| *VIF [23]* | 1.1745 |
| *FSIM [7]* | 0.5269 |
| *IW-SSIM [16]* | 0.5196 |
| *MS-SSIM [17]* | 0.1379 |
| *GS [15]* | 0.0899 |
| *GSD [5]* | 0.0481 |
| *SSIM [8]* | 0.0388 |
| *G-SSIM [6]* | 0.0379 |
| *GMSD* | 0.0110 |
| *GMSM* | 0.0079 |
| *PSNR* | 0.0016 |

Table IV shows the running time of the 13 IQA models on an image of size 512×512. All algorithms were run on a ThinkPad T420S notebook with Intel Core i7-2600M CPU@2.7GHz and 4G RAM. The software platform used to run all algorithms was MATLAB R2010a (7.10). Apart from G-SSIM and GSD, the MATLAB source codes of all the other methods were obtained from the original authors. (It should be noted that whether the code is optimized may affect the running time of an algorithm.) Clearly, PSNR is the fastest, followed by GMSM and GMSD. Specifically, it costs only 0.0110 second for GMSD to process an image of size 512×512, which is 3.5 times faster than SSIM, 47.9 times faster than FSIM, and 106.7 times faster than VIF.

### G. Discussions

Apart from being used purely for quality assessment tasks, it is expected that an IQA algorithm can be more pervasively used in many other applications. According to [1], the most common applications of IQA algorithms can be categorized as follows: 1) quality monitoring; 2) performance evaluation; 3) system optimization; and 4) perceptual fidelity criteria on visual signals. Quality monitoring is usually conducted by using no reference IQA models, while FR-IQA models can be applied to the other three categories. Certainly, SSIM proved to be a milestone in the development of FR-IQA models. It has been widely and successfully used in the performance evaluation of many image processing systems and algorithms, such as image compression, restoration and communication, etc. Apart from performance evaluation, thus far, SSIM is not yet pervasively used in other applications. The reason may be two-fold, as discussed below. The proposed GMSD model might alleviate these problems associated with SSIM, and has potentials to be more pervasively used in a wider variety of image processing applications.

First, SSIM is difficult to optimize when it is used as a fidelity criterion on visual signals. This largely restricts its applications in designing image processing algorithms such as image compression and restoration. Recently, some works [36-38] have been reported to adopt SSIM for image/video perceptual compression. However, these methods are not “one-pass” and they have high complexity. Compared with SSIM, the formulation of GMSD is much simpler. The calculation is mainly on the gradient magnitude maps of reference and distorted image, and the correlation of the two maps. GMSD can be more easily optimized than SSIM, and it has greater potentials to be adopted as a fidelity criterion for designing perceptual image compression and restoration algorithms, as well as for optimizing network coding and resource allocation problems.

Second, the time and memory complexity of SSIM is relatively high, restricting its use in applications where low-cost and real-time implementation is required. GMSD is much faster and more scalable than SSIM, and it can be easily adopted for tasks such as real time performance evaluation, system optimization, etc. Considering that mobile and portable devices are becoming much more popular, the merits of simplicity, low complexity and high accuracy of GMSD make it very attractive and competitive for mobile applications.

In addition, it should be noted that with the rapid development of digital image acquisition and display technologies, and the increasing popularity of mobile devices and websites such as YouTube and Facebook, current IQA databases may not fully represent the way that human subjects view digital images. On the other hand, the current databases, including the three largest ones TID2008, LIVE and CSIQ, mainly focus on a few classical distortion types, and the images therein undergo only a single type of distortion. Therefore, there is a demand to establish new IQA databases, which should contain images with multiple types of distortions [40], images collected from mobile devices [41], and images of high definition.

## IV. Conclusion

The usefulness and effectiveness of image gradient for full reference image quality assessment (FR-IQA) were studied in this paper. We devised a simple FR-IQA model called gradient magnitude similarity deviation (GMSD), where the pixel-wise gradient magnitude similarity (GMS) is used to capture image local quality, and the standard deviation of the overall GMS map is computed as the final image quality index. Such a standard deviation based pooling strategy is based on the consideration that the variation of local quality, which arises from the diversity of image local structures, is highly relevant to subjective image quality. Compared with state-of-the-art FR-IQA models, the proposed GMSD model performs better in terms of both accuracy and efficiency, making GMSD an ideal choice for high performance IQA applications.

## Acknowledgment

This work is supported by Natural Science Foundation of China (No. 90920003 and No. 61172163) and HK RGC General Research Fund (PolyU 5315/12E).